\documentclass[journal]{IEEEtran}

\usepackage{amsmath,graphicx,amsmath}
\usepackage{amsfonts}
\usepackage{color}
\usepackage{amssymb}
\usepackage{amsthm}

\ifCLASSINFOpdf
\else
   \usepackage[dvips]{graphicx}
\fi
\usepackage{url}

\usepackage{graphicx}

\begin{document}

\title{Video object tracking based on YOLOv7 and DeepSORT}

\author{Feng Yang, \IEEEmembership{Member, IEEE}, Xingle Zhang and Bo Liu
\thanks{F. Yang, X. Zhang and B. Liu are all with the Key Lab of Information Fusion Technology (Ministry of Education), School of Automation, Northwestern Polytechnical University, Xi’an 710129, China, e-mail: yangfeng@nwpu.edu.cn,zxlyyy@mail.nwpu.edu.cn,503790475@qq.com}}

\markboth{Journal of \LaTeX\ Class Files, Vol. 14, No. 8, August 2015}
{Shell \MakeLowercase{\textit{et al.}}: Bare Demo of IEEEtran.cls for IEEE Journals}
\maketitle

\begin{abstract}
Multiple object tracking (MOT) is an important technology in the field of computer vision, which is widely used in automatic driving, intelligent monitoring, behavior recognition and other directions. Among the current popular MOT methods based on deep learning, Detection Based Tracking (DBT) is the most widely used in industry, and the performance of them depend on their object detection network. At present, the DBT algorithm with good performance and the most widely used is YOLOv5-DeepSORT. Inspired by YOLOv5-DeepSORT, with the proposal of YOLOv7 network, which performs better in object detection, we apply YOLOv7 as the object detection part to the DeepSORT, and propose YOLOv7-DeepSORT. After experimental evaluation, compared with the previous YOLOv5-DeepSORT, YOLOv7-DeepSORT performances better in tracking accuracy.
\end{abstract}

\begin{IEEEkeywords}
Multiple Object Tracking;
Object Detection;
DeepSORT;
YOLO;
\end{IEEEkeywords}

\IEEEpeerreviewmaketitle

\section{Introduction}

Multiple object tracking (MOT) generally refers to the detection and ID tracking of multiple targets in the video, such as pedestrians, cars, animals, etc., without knowing the number of targets in advance. Different targets have different IDs in order to achieve subsequent trajectory prediction, accurate search and other work. MOT is an important technology in the field of computer vision, which is widely used in automatic driving, intelligent monitoring, behavior recognition and other directions. In MOT, we must not only face the challenges of occlusion, deformation, motion blur, crowded scene, fast motion, illumination change, scale change and so on in single object tracking, but also face complex problems such as trajectory initialization and termination, mutual interference between similar targets and so on. Therefore, MOT is still a very challenging direction in image processing, which has attracted the long-term investment of many researchers.

Visual object tracking has not developed for a long time, mainly in the past ten years. The early classical methods include Meanshift \cite{1Cheng} and particle filter \cite{2Okuma}, but the overall accuracy of these algorithms is low, and they are mainly single object tracking, which is difficult to meet the requirements of complex scenes. In recent years, with the rapid development of deep learning, the performance of target detection has been improved by leaps and bounds, and the scheme of detection based tracking (DBT) has also been born. It has quickly become the mainstream framework of current MOT, which greatly promotes the progress of MOT tasks. At the same time, recently, there have been joint frameworks based on detection and tracking and frameworks based on attention mechanism, which have begun to attract researchers' attention.

\section{Related works}

The current MOT framework can be divided into three types: MOT based on tracking by detection (DBT), MOT based on joint detection and tracking, and MOT based on attention mechanism. The first one is more widely used in industry.

The process of DBT framework is: First, detect the targets in each frame of the video sequence, cut the targets according to the bounding box, and get all the targets in the image. Then, it is transformed into the problem of target correlation between the front and back frames. The similarity matrix is constructed through IoU, appearance feature, etc., and solved by Hungarian algorithm, greedy algorithm, etc. The tracking effect of this kind of algorithm depends on the performance of its object detection network. At present, the most used detection network is the YOLO series network, such as YOLOv3 \cite{3Redmon}, YOLOv4 \cite{4Bochkovskiy}, YOLOv5 \cite{5Jocher}. SORT \cite{6Bewley} and DeepSORT \cite{7Wojke} are the most concerned tracking algorithms in the industry. The core of SORT is Kalman filter and Hungarian matching. The position of the target is predicted by Kalman filter, and the prediction result of the object detection network like YOLO is matched with the result of Kalman filter by Hungarian matching. SORT is a practical MOT algorithm. However, due to the variable target motion and frequent occlusion in reality, the algorithm has a high number of identity switches. Therefore, the author adds cascade matching and other functions on its basis, and proposes DeepSORT with better performance.

MOT based on joint of detection and tracking combines detection and tracking framework. This kind of algorithm generally detects the two adjacent frames of the video, and then uses different strategies to judge the similarity of the targets existing in the two frames at the same time, so as to track and predict. Typical algorithms include D\&T \cite{8Feichtenhofer}, MOTDT \cite{9Chen}, FairMOT \cite{10Zhang}, CenterTrack \cite{11Zhou}, etc.

MOT based on attention mechanism is to apply Transformer \cite{12Vaswani} to MOT. At present, there are mainly TransTrack \cite{13Sun} and TrackFormer \cite{14Meinhardt}. TransTrack takes the feature map of the current frame as the Key, and takes the target feature Query of the previous frame and a group of target feature Query learned from the current frame as the input Query of the whole network.

\section{YOLOv7-DeepSORT}

\subsection{YOLOv7}

YOLOv7 \cite{15Wang} is the latest work of YOLO series. This network further improves the detection speed and accuracy on the basis of the previous work. Specifically, in terms of the overall architecture, the paper proposes E-ELAN, uses expand, shuffle, merge cardinality to achieve the ability to continuously enhance the learning ability of the network without destroying the original gradient path. E-ELAN can guide different groups of computational blocks to learn diverse features. The paper also proposes a compound model scaling method to maintain the properties that the model had at the initial design and maintains the optimal structure.

In terms of network optimization strategy, the paper introduces model re-parameterization and dynamic label assignment, analyzes their existing problems, and improves them. For the former, the author believes that because RepConv \cite{16Ding} has identity connection, direct access to the cascade of ResNet \cite{17He} or DenseNet \cite{18Huang} will provide more gradient diversity for different characteristic graphs, thus destroying the network structure. Therefore, the author removed the identity connection in RepConv and designed the planned re-parameterized convolution, realizing the efficient combination of re-parameterized convolution and different networks. For the latter, the paper uses the idea of Deep supervision \cite{19Lee} and adds an additional auxiliary head structure in the middle layer of the network as an auxiliary loss to guide the weight of the shallow network. A new label assignment method is designed for this structure.

\subsection{DeepSORT}

SORT algorithm uses a simple Kalman filter to deal with the correlation of frame-by-frame data, and uses the Hungarian algorithm to measure the correlation. This algorithm has achieved good performance at high frame rate. However, since SORT ignores the appearance feature of the detected target, it will be accurate only when the uncertainty of target state estimation is low. In addition, in order to improve the tracking efficiency, SORT deletes the target that has not been matched in a continuous frame, but this causes the problem of ID switch, that is, the ID assigned to the target is easy to change constantly.

Therefore, DeepSORT adds appearance information and borrows ReID model to extract appearance features, reducing the number of ID switches by 45\%. DeepSORT also turns SORT's matching mechanism based on IoU cost matrix into a mechanism of Matching Cascade and IoU matching. Specifically, the core idea of Matching Cascade is to give greater priority to track matching to the targets that appear more frequently in the long-term occluded targets. This method solves the matching problem of targets that have been occluded for a long time. DeepSORT performs IoU matching on unmatched tracks and detection targets in the final stage of matching, which can alleviate large changes caused by apparent mutations or partial occlusion. In addition, DeepSORT borrows the ReID model to require a well distinguishing feature embedding from the output of the object detection network for calculating the similarity.

\subsection{YOLOv7-DeepSORT}

Considering the excellent performance of YOLOv7 in object detection tasks, we refer to YOLOv5-DeepSORT \cite{20Mikel} and replace YOLOv7 with the object detection model of the network to obtain YOLOv7-DeepSORT. The operation process of YOLOv7-DeepSORT is shown in Fig.1. The network trains YOLOv7 and ReID separately.

\begin{figure}[htb]
	\centering
	\includegraphics[width=3in]{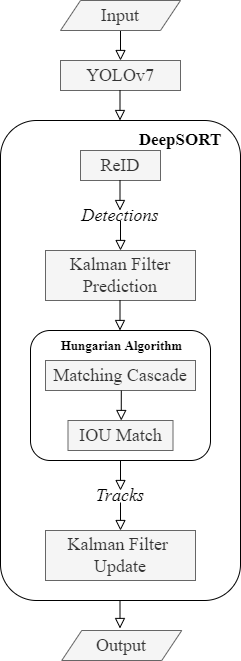}
	\caption{The operation process of YOLOv7-DeepSORT. The input is sent to YOLOv7 to detect the targets. Then, ReID model requires a well distinguishing feature embedding from the output of YOLOv7. Kalman filter predicts the trajectory of each target. And the outputs of YOLOv7 and Kalman filter are sent to the Hungarian algorithm for matching to obtain the trajectory of the target.}
	\label{Fig3Fusion}
\end{figure}

\section{Experiments}

In the experimental part, we evaluated the performance of YOLOv7-DeepSORT on the 02, 04, 05, 09, 10, 11, 13 sequences of MOT16 \cite{21Milan} challenge and compared it with YOLOv5-DeepSORT. The parameters of the two in the DeepSORT part were set to be exactly the same. Among them, the models used in the object detection part are YOLOv7, YOLOv5s, YOLOv5m and YOLOv5l, all of which use the official pre-trained model (the version of YOLOv5 is r6.1). The experiment was completed on GTX 3080Ti.

The evaluation metrics used in the experiment are as follows:

MOTA \cite{22Bernardin}: Multi-Object Tracking Accuracy. This measure combines three error sources: false positives, missed targets and identity switches.

MOTP \cite{22Bernardin}: Multi-Object Tracking Precision: Summary of overall tracking precision in terms of bounding box overlap between ground-truth and reported location.

IDF1 \cite{23Ristani}: ID F1 Score. The ratio of correctly identified detections over the average number of ground-truth and computed detections.

IDs: Number of Identity Switches. 

ML: Mostly Lost Targets. The ratio of ground-truth trajectories that are covered by a track hypothesis for at most 20\% of their respective life span.

MT: Mostly Tracked Targets. The ratio of ground-truth trajectories that are covered by a track hypothesis for at least 80\% of their respective life span.

FP: The total number of false positives.

FN: The total number of false negatives (missed targets).

The results of the experiment are shown in Table 1.

It can be seen from Table 1. that YOLOv7-DeepSORT (hereinafter referred to as YOLOv7) is higher than YOLOv5-DeepSORT (hereinafter referred to as YOLOv5s/m/l) in MOTA, MOTP and IDF1, so its tracking accuracy is indeed better. For ID switch, YOLOv7 is better than YOLOv5l. For ML and MT, YOLOv7 is slightly worse than YOLOv5l, and better than YOLOv5s and YOLOv5m, and has a good tracking effect on the target. In general, YOLOv7-DeepSORT has significantly improved the tracking accuracy compared with YOLOv5-DeepSORT.

\begin{table}[htp]	
	\centering
	\caption{Tracking results on the MOT16 challenge. We compare the tracking performance of YOLOv7-DeepSORT and YOLOv5(s/m/l)-DeepSORT.}
	\begin{tabular}{|c|c|c|c|c|} 
		\hline 
		Model &  YOLOv5s & YOLOv5m & YOLOv5l & YOLOv7 \\ 
		\hline 
		MOTA &  39.60 & 39.01 & 40.77 & \textcolor{red}{40.82}\\ 
		\hline
		MOTP &  80.85 & 81.87 & 81.96 & \textcolor{red}{82.01}\\ 
		\hline
		IDF1 &  52.39 & 51.56 & 52.43 & \textcolor{red}{53.65}\\ 
		\hline
		IDs &  \textcolor{red}{432} & \textcolor{red}{432} & 547 & 514\\ 
		\hline
		ML &  39.65\% & 33.27\% & \textcolor{red}{31.92\%} & 32.11\%\\ 
		\hline
		MT & 15.45\% & 17.41\% & \textcolor{red}{20.70\% }& 20.12\%\\ 
		\hline
		FP &  \textcolor{red}{5375} & 7612 & 7853 & 7940\\ 
		\hline
		FN &  60882 & 59297 & \textcolor{red}{56990 }& 57434\\ 
		\hline
	\end{tabular}	
	\label{Tab01}
\end{table}

\section{Conclusion}

We add YOLOv7 as object detection network to DeepSORT and get YOLOv7-DeepSORT. Experiments show that this network has better tracking accuracy than YOLOv5-DeepSORT. Thanks to the excellent generalization of YOLOv7 and DeepSORT, the YOLOv7-DeepSORT is also applicable to all kinds of target tracking tasks.

%
%


\end{document}